\documentclass[11pt]{article}
\usepackage{times}
\usepackage{url}
\usepackage{amssymb,txfonts}
\begin{document}

\title{Propositional Interpretability in Artificial Intelligence}

\author{David J. Chalmers}
\date{}
\maketitle

\abstract{Mechanistic interpretability is the program of explaining what AI systems are doing in terms of their internal mechanisms. I analyze some aspects of the program, along with setting out some concrete challenges and assessing progress to date. I argue for the importance of propositional interpretability, which involves interpreting a system's mechanisms and behavior in terms of propositional attitudes: attitudes (such as belief, desire, or subjective probability) to propositions (e.g. the proposition that it is hot outside). Propositional attitudes are the central way that we interpret and explain human beings and they are likely to be central in AI too. A central challenge is what I call thought logging: creating systems that log all of the relevant propositional attitudes in an AI system over time. I examine currently popular methods of interpretability (such as probing, sparse auto-encoders, and chain of thought methods) as well as philosophical methods of interpretation (including those grounded in psychosemantics) to assess their strengths and weaknesses as methods of propositional interpretability.}

\section{Introduction}

Mechanistic interpretability is one of the most exciting and important research programs in current AI. As I understand it, {\em interpretability} (in a broad sense) is the practice of explaining what AI systems are doing in terms a human can understand.  Mechanistic interpretability focuses on explaining what AI systems are doing in terms of their internal mechanisms.\protect\footnotemark\ 

\footnotetext{Thanks to audiences at the British Academy and at the NYU Center
for Data Science in October 2024, and to Tal Linzen and Nick Shea.
This article will be the basis for an invited talk at the AAAI
conference in March 2025.  For now this is a rough draft.  For a short
version of this paper, oriented more to AI than philosophy, read
sections 1, 5, and 7.}

Mechanistic interpretability is important for a number of reasons.  In {\em AI safety}, mechanistic interpretability might help us to know an AI system's goals and plans by examining its internal processes.  In {\em AI ethics}, mechanistic interpretability might allow us to better understand an AI system's reasons for its decisions and the biases that enter into them.  In {\em AI cognitive science}, mechanistic interpretability offers the prospect of a scientific explanation of AI systems, akin to the scientific explanations we aim for with human beings.

I will argue for the importance of a special sort of interpretability, which I call {\em propositional interpretability}.  This involves interpreting a system's mechanisms and behavior in terms of {\em propositional attitudes} (or generalizations thereof).  In ordinary human psychology, propositional attitudes are attitudes (such as belief, desire, or subjective probability) to propositions (e.g. the proposition {\em It is hot outside}).  Someone might believe that it is hot outside, or desire that it is hot outside, or have subjective probability 0.5 that it is hot outside.  These are very different states and play a very different role in predicting and explaining action.

Propositional interpretability is crucially important in AI.  For example, in AI safety, it is crucial to know an AI system's goals (desires) and its models of the world (beliefs).  To know these, we need to know not just the concepts or features that are active in the system; we need to know the system's propositional attitudes.  It is very different for an AI system to believe that a certain dangerous outcome (an earthquake or a war, say) has occurred, and for it to have that outcome as a goal.  The same goes in AI ethics.  It is one thing for an AI system to believe that members of a certain demographic group are often denied loans, and another thing for the system to have this as a goal. 

Propositional interpretability lies at a nexus between AI, cognitive
science, and philosophy.  In this area, philosophy and cognitive
science can contribute to AI by providing foundations for
interpretability and tools to understand AI systems better.\protect\footnotemark\   AI may
also contribute to philosophy and cognitive science, by offering new
insights into propositional attitudes and their role in action and
communication.

\footnotetext{For discussions of how cognitive neuroscience may offer tools for AI
interpretability, see He et al (2024), Lindsay and Bau (2023), and
Vilas et al (2024).}

As we will see, the research program of propositional interpretability is already an active program, albeit under many different names, and without much focus on attitudes other than (sometimes) belief and knowledge.\protect\footnotemark\   My aim is to help clarify the program and build some philosophical foundations for it (in sections 2-4 and 6), along with setting out some concrete challenges (section 5), assessing methods and progress to date (section 7), and answering objections (section 8).

\footnotetext{For existing work on propositional interpretability (under various labels), see e.g. B. Li et al 2021; Meng et al 2022a; K. Li et al 2023; Feng et al 2024.}

My central challenge for propositional interpretability is what I call {\em thought logging}.  The aim is to create systems that log all of the (most relevant) propositional attitudes in an AI system over time.  This is a long-term project but at least partial progress should be possible.  I will examine existing methods of interpretability (such as probing, sparse auto-encoders, and chain of thought methods) to assess their potential strengths and weaknesses as methods of thought logging.

\section{Varieties of interpretability}

I'll start by breaking down the varieties of interpretability in a fine-grained way, to clarify the area and to make clearer where propositional interpretability fits in.\protect\footnotemark\ 

\footnotetext{For foundational and philosophical discussions of interpretability, including numerous relevant conceptual and terminological distinctions, see Buckner and Milliere (forthcoming), Doshi-Velez and Kim (2017), Erasmus et al (2021), Grzankowski (forthcoming), Krishnan (2019), Lipton (2018), and Saphra and Wiegresse (2024).}

Starting at the beginning: we might say that to {\em interpret} an AI system is to explain what the system is doing in human-understandable terms.  {\em Interpretability} is a name for the practice of interpreting AI systems.\protect\footnotemark\ 

\footnotetext{For this sort of definition of ``interpret'', see Doshi-Velez and Kim
2017.  The practice of interpreting AI systems might more simply be
called ``interpretation'', but the label ``interpretability'' (which may
have originated as a name for the somewhat different practice of
designing AI systems that are easily interpretable) has stuck.}

I have defined interpretability in terms of explanation, and the terms {\em interpretability} and {\em explainability} are sometimes used interchangeably.  But it is also common for the terms to be used differently. There are various ways to understand the difference, but perhaps the core difference for my purposes stems from a difference in who a ``human-understandable'' explanation is directed at.

{\em Explainability} is explanation for ordinary humans.  It is especially directed at end users and others affected by an AI system's decisions.  For example, explainability might involve explaining to a doctor or a patient why the patient was diagnosed with a given disease. Explainability typically involves explanations in nontechnical terms.

{\em Interpretability} (in the narrower sense) is explanation for theorists.  It is especially directed at scientists and AI researchers who are trying to understand what an AI system is doing.  For example, interpretability might involve explaining how a language model learns from data and how it encodes a model of the world.  While explainability is typically nontechnical, interpretability is frequently quite technical.

Both of these projects are important, but my focus here is on
interpretability, not explainability.  Interpretability divides in
turn into at least two classes.  {\em Behavioral} interpretability
analyzes an AI system's input/output behavior to understand what the
system is doing.  {\em Mechanistic} interpretability analyzes an AI
systems internal mechanisms to help explain (for theorists) what the
system is doing.\protect\footnotemark\ 

\footnotetext{Behavioral and mechanistic interpretability are also sometimes known as 
black-box and white-box interpretability (respectively), or as outer and inner interpretability (see Grzankowski forthcoming for discussion).}

I'll focus mainly on mechanistic interpretability, which can itself be
subdivided.  {\em Algorithmic} interpretability aims to understand the
underlying algorithms that an AI system is executing, in clear
theoretical terms.  One major line of research on algorithmic
interpretability focuses on circuits, which are distinctive
algorithmic processes that can be found within a given AI system: for
example, the distinctive circuits that can be formed by chaining
together multiple attention heads in a transformer.

{\em Representational} interpretability aims to understand the internal {\em representations} that an AI system is using.  Here representations are not just any internal pattern of activity, as one common usage in AI has it.  Instead, representations are internal entities (e.g. symbols or patterns of activity) that represent or stand for other entities in the world.  For example, the symbol ``2'' represents the number two.  Certain patterns of activity in a neural network might represent Paris, or cats, or happiness.\protect\footnotemark\ 

\footnotetext{``Mechanistic interpretability'' originated as a label for algorithmic interpretability in circuits (see the online ``Circuits'' thread and Olah et al 2020), where this contrasted with then-current work in representational interpretability including salience maps.  These days the term is often used more inclusively for any interpretability work involving internal mechanisms, including work on representational interpretability.  (For example, the work on sparse-auto encoders that I discuss in section 7 focuses on representations more than algorithms but is classed as mechanistic interpretability all the same.)  There is also some work on representational interpretability (e.g. using representation theorems from decision theory) that does not analyze inner mechanisms and so falls under the heading of behavioral interpretability rather than mechanistic interpretability.}

My focus will be especially on representational interpretability, not least because this sort of interpretability has such a clear role in AI safety, AI ethics, and AI cognitive science.  Representational interpretability itself
divides into at least two classes.

{\em Conceptual interpretability} aims to understand the {\em concepts} that an AI system is using: concepts such as {\em cat}, {\em plus}, {\em election}, and {\em Golden Gate Bridge}.  Concepts often correspond to entities in the world, or to categories thereof.

{\em Propositional interpretability}, as we've already seen, aims to understand the propositional attitudes that an AI system is using: e.g. believing {\em the Golden Gate Bridge is large}.  Here, propositional attitudes are attitudes (like belief, desire, or subjective probability) to propositions (like {\em 2+2=4}).  I'll discuss these further in the next section.\protect\footnotemark\ 

\section{Propositional attitudes}

What are propositions, and what are propositional attitudes?

On one standard view, propositions are structured entities that are composed of concepts: for example, the proposition {\em the Golden Gate Bridge is large} is composed of concepts such as {\em Golden Gate bridge} and {\em red}.  (I'll consider other views, such as the view that propositions are sets of possible worlds, shortly.) It is widely held that sentences in natural languages (or at least utterances thereof), such as ``2+2=4'' or ``The Golden Gate Bridge is large'', express propositions.  Propositions can be evaluated for truth or falsity, and serve as truth-conditions for sentences: for example, the sentence ``2+2=4'' is true if the corresponding proposition {\em 2+2=4} is true.

What are propositional attitudes?  Perhaps the canonical propositional attitudes are {\em belief} and {\em desire}.  When I believe that 2+2=4, I stand in the attitude of belief to the proposition {\em 2+2=4}. When I want Australia to win the Ashes, I stand in the attitude of desire to the proposition that {\em Australia wins the Ashes}.

Another important propositional attitude is {\em credence}, which is the
numerical counterpart of belief. Credence is subjective probability,
or degree of belief: if I have credence 0.25 in {\em It will rain today},
that's to say that my subjective probability or my degree of belief
that it will rain today is one in four.  One can also postulate a
numerical correlate of desire, sometimes called {\em utility}. Very
roughly speaking, {\em It will rain today} has higher utility for me than
{\em It will not rain today} if I prefer the former to the latter. Unlike
the case of credence, there is no obvious absolute numerical scale for
degree of desire, so it is arguable that utility really involves a
relative scale, one that could be cashed out in terms of the attitude
of {\em preference}.

Two other important propositional attitudes are {\em intention} and
{\em supposition}.  Intention is intending to do something: when I intend
to eat lunch soon, I stand in the relation of intending to {\em I eat
lunch soon}.  Supposition is supposing that something in the case,
possibly for conditional (``If ... then ...'') reasoning.  For example, I
might suppose {\em It is raining today} to reach the conclusion {\em If it is
raining today, the match will be cancelled}.  Other propositional
attitudes include {\em hope}, {\em fear}, and many more: I can hope Australia
wins the Ashes, I can fear they won't, and so on.

Propositional attitudes can be divided into {\em dispositional} and
{\em occurrent}. Roughly speaking, occurrent attitudes are those that are active at a given time.  (In a neural network, these would be encoded in neural activations.)  Dispositional attitudes are typically inactive but can be activated. (In a neural network, these would be encoded in the weights.)  For example, I believe {\em Paris is the capital of France} even when I am asleep and the belief is not active.  That is a dispositional belief.  On the other hand, I may actively judge {\em France has won more medals than Australia}.  That is an occurrent mental state, sometimes described as an ``occurrent belief'', or perhaps better, as a ``judgment'' (so judgments are active where beliefs are dispositional).  One can make a similar distinction for desires and other attitudes.

Propositional attitudes are absolutely central to how we understand, explain, and predict what other human beings do.  We understand them as having desires or preferences (for food, for love, for success), and as performing actions that might lead to those desires being satisfied, at least if their beliefs about the world are true.  Perhaps all this can be understood more scientifically by invoking credences and utilities rather than beliefs and desires.  But it is hard to understand human action without a framework like this.

Propositional attitudes are also likely to be central to the human understanding of AI systems.\protect\footnotemark\   It is hard to understand an AI system without understanding its goals (which correspond to desires) and its models of the world (which correspond to beliefs).  We also often need to understand an AI system's probabilities (which correspond to credences), which play a central role in many AI systems.  Like humans and other organisms, AI systems perform actions to try to satisfy their goals in light of these probabilities and in light of their models of the world.

\footnotetext{For recent philosophical discussions of whether language models have propositional attitudes, see Goldstein and Levinstein (2024), Lederman and Mahowald (2024), and Shanahan (2022), as well as a classic discussion by Ramsey, Stich, and Garon (1989).  For discussions of belief in language models, see Herrmann and Levinstein (forthcoming), Levinstein and Herrmann (2024), and Schwitzgebel (2023).  For knowledge, see Yildirim and Paul (2024).  For credence, see Keeling and Street (2025).}

To understand AI systems in representational terms, we need propositional interpretability.  Conceptual interpretability is not enough.  We need propositions, not just concepts.  Even if the concepts {\em kill} and {\em humans} are active, the system could be representing {\em Kill humans} or {\em Don’t kill humans}---a crucial distinction.  Likewise, we need attitudes, not just propositions.  Even if {\em Australians are unsuccessful} (my attempt at a negative evaluation of a demographic group) is represented, this could be a desire (goal), a belief (model), a credence (probability), or a supposition (if ... then ...), with very different results.

Perhaps it is less natural to apply terms such as ``belief'', ``desire'', and ``credence'' to AI systems than to humans, since these terms have many loaded connotations from the human case.  But less loaded counterparts of these terms are common when describing AI systems: for example, ``models'', ``goals'', and ``probabilities'', as above.

It is entirely possible that these familiar propositional attitudes
will need to be refined to give powerful explanation of AI systems.
As Patricia and Paul Churchland have stressed, belief and desire are
crude terms from folk psychology, inheriting much of the vagueness and
ambiguity of ordinary language.  Even probabilities, models, and goals
might be replaced by more precise categories.  In effect,
propositional interpretability for AI may involve a project of
{\em conceptual engineering}, where we engineer new and refined categories
of propositional attitudes over time.

This conceptual engineering project may involve a move from familiar
propositional attitudes to {\em generalized propositional attitudes}.
These may involve generalized attitudes that go beyond the folk
categories of belief, desire, and so on.  They may also involve
generalized objects of these attitudes that go beyond traditional
propositions.

Generalized propositional attitudes also have the advantage of
covering many sorts of representations that are not always counted as
propositional attitudes in the traditional sense.  For example,
pictorial or map-like representations in the brain are not naturally
understood as involving attitudes to traditional propositions that are
structured roughly like sentences.  But these representations still
say something about the world.  What they say about the world can be
true or false.  As philosophers put it, they have contents that can be
specified in some other form: as sets of possible worlds, for example,
or perhaps as spatial structures.  When a mouse has a map-like
representation of places, or when a bee senses its environment, it can
be seen as endorsing a structure of this sort.  I count will count
even these non-sentential states as generalized propositional
attitudes.\protect\footnotemark\ 

\footnotetext{Grzankowski and Montague (2018) is a collection of articles on non-propositional forms of intentionality.  Many of these can be seen as involving generalized propositional attitudes, some with attitudes to non-sententially-structured propositions, and some with attitudes to related entities that are not propositions (for example, intention can be construed as an attitude to an action, while worship can be construed as an attitude to an object).  It is perhaps a stretch to use ``generalized propositional attitude'' to include the last category (attitudes to non-propositions), but there's isn't an obviously better term.}

Another advantage of invoking generalized propositional attitudes
(discussed further in section 8) is that it allows us to sidestep some
debates about whether AI systems have minds.  Beliefs and desires are
usually understood as mental states.  If so, then only systems with
minds can believe something.  It is highly controversial whether AI
systems have minds, so it is controversial whether they believe
anything.  By contrast, it is somewhat less controversial to say that
AI systems can have models or goals, because these are not usually
understood as mental states that require a mind.

To take a very simple system: a thermostat may not have propositional
attitudes in the ordinary sense, but it plausibly has representations
(e.g. representing {\em the temperature is now 60 degrees}) and goals
(e.g. aiming at {\em the temperature is 70 degrees}).  These can be
naturally understood as generalized propositional attitudes.  Of
course it takes some work to precisely define the conditions for these
or any other propositional attitudes, and it takes some work to
analyze just what role these generalized propositional attitudes can
play in explaining action, but this is all part of the project of
propositional interpretability.\protect\footnotemark\ 

\footnotetext{Schwitzgebel (2023) offers ``belief*'', a conceptually engineered version of belief, in order to sidestep questions about consciousness.}

\section{Radical interpretation}

These interpretability projects all have analogs in the human case.  Humans have been interpreting other humans since the human species began.  Ordinary people use patterns in behavior to explain what other people are doing, most often in propositional terms. Scientists and philosophers have used theoretical tools to explain human behavior for at least thousands of years.  More recently, they have used tools involving internal mechanisms to do the same thing.

There is a well-known philosophical program for interpreting human beings in propositional terms. The program of ``radical interpretation'' was set out in the 1970s by two of the leading analytic philosophers of the post-war period.  Donald Davidson set out the name and the original program in a 1973 article. David Lewis gave a canonical statement of the program (or at least one version of it) in a 1974 reply to Davidson, which starts as follows:

\begin{quote} ``Imagine that we have undertaken the task of coming to know Karl as a person. We would like to know what he believes, what he desires, what he means, and anything else about him that can be explained in terms of these things. We seek a two-fold interpretation: of Karl's language, and of Karl himself. And we want to know his beliefs and desires in two different ways. We want to know their content as Karl could express it in his own language, and also as we could express it in our language. ... 

Imagine also that we must start from scratch. At the outset we know nothing about Karl's beliefs, desires, and meanings. Whatever we may know about persons in general, our knowledge of Karl in particular is limited to our knowledge of him as a physical system. But at least we have plenty of that knowledge---in fact, we have all that we could possibly use. Now, how can we get from that knowledge to the knowledge we want?

I can diagram the problem of radical interpretation as follows. Given P, the facts about Karl as a physical system, solve for the rest. \end{quote}

Here, ``the rest'' refers to Karl's beliefs, desires, and meanings. So Lewis's statement of the problem amounts to: given the physical facts about a system, solve for the system's beliefs, desires, and meanings.  For my purposes, the focus on beliefs and desires is especially important.  Since these are propositional attitudes, Lewis's project is a version of propositional interpretability.

Davidson's version of radical interpretation differs in an important
way.  Davidson's version in effect says: given the {\em behavioral} facts
about a system, solve for its beliefs, its desires, and its meanings.
Davidson's interpreter is restricted to behavior and only behavior.
As such, his is a sort of behavioral interpretion, falling into a long
program of broadly behaviorist approaches interpreting human beings.
It was preceded by W.V. Quine's (1960) program of ``radical
translation'', and succeeded by Daniel Dennett's (1987) project of
understanding human minds through the ``intentional stance''.

These behaviorist programs have gradually fallen out of favor,
replaced by programs such as Lewis's, which involve what we might call
{\em physical interpretation}: using physical facts to solve for
propositional attitudes.\protect\footnotemark\   Unlike Davidson's program, Lewis's physical
interpretation is consistent with mechanistic interpretability, where
internal mechanisms may play a key role in interpretation.

The advent of AI interpretability has in effect introduced a third
project here, which we might call {\em computational} interpretation:
given the computational facts about a computational system, such as
the algorithm it is running plus the computational states it is in
(plus relevant environmental facts), solve for its propositional
attitudes.  In the case of an artificial neural network, the
computational facts will include its structure, weights, activations,
inputs and outputs, and history.  The project of propositional
interpretability for AI systems involvings moving from computational
facts (plus relevant environmental facts) to propositional attitudes.\protect\footnotemark\ 

\footnotetext{A paragraph on radical vs nonradical interpretation is
needed here.}

\section{Thought Logging}

A concrete challenge for research in propositional interpretability is
to construct a {\em thought logging} system: a system that logs all (or as
many as possible) of an AI system's propositional attitudes over time.
A thought logging system is a meta-system that takes a specification
of the algorithmic facts about an AI system as input (perhaps along
with relevant environmental facts) and produces a list of the system's
current and ongoing propositional attitudes as outputs.

A log (in an ultra-simple form) might look something like this:

\begin{quote}
Goal: I win this game of chess.

Judge (credence 0.8): If I move Qf8, I will win

Goal: I move Qf8.

Action: I move Qf8.
\end{quote}

Now, it is likely that a given AI system may have an infinite number
of propositional attitudes, in which case a full log will be
impossible.  For example, if a system believes a proposition p, it
arguably dispositionally believes p-or-q for all q.  One could perhaps
narrow down to a finite list by restricting the log to occurrent
propositional attitudes, such as active judgments.  Alternatively, we
could require the system to log the most significant propositional
attitudes on some scale, or to use a search/query process to log all
propositional attitudes that meet a certain criterion.

Interestingly, in ``Radical Interpretation'', Lewis offered a possible
format for entries in a thought log.

\begin{quote}
Karl {believes/desires}, to degree d, at time t, the proposition expressed, in context c, by the sentence `——' of {our/Karl’s} language.
\end{quote}

Lewis restricted himself to belief and desire, but allowed degrees of beliefs (credences) and degrees of desire(utilities).  He says that he hopes that all propositional attitudes can be analyzed in these terms, but if not, then other attitudes should be included.  In principle, I think we should be open to including other attitudes in the class of generalized propositional attitudes.

Lewis also restricts himself to propositions expressed by sentences in or language or the subject's language.  Sentences in our language are clearly useful for interpretability, whereas sentences in a different language may require translation to be useful. We also should be open to propositions that are not expressible in our language or the subject's language.  To express these in a log form, we may need new expressive resources, such as new notation, in order to capture these propositions as well as possible (even if still imperfectly).

An ideal form of thought logging would include various extensions.
{\em Reason logging} would display a system's reasons for holding a given
propositional attitude wherever possible, possibly via support links
from earlier attitudes to later attitudes whenever the former plays a
substantial role in the formation of the latter.  {\em Mechanism logging}
could enhance thought logging with an indication of the internal
mechanisms responsible for any given propositional attitudes, whenever
possible.  Reason logging may help a great deal with interpretability
by ordinary humans, while mechanism logging may help with scientific
and mechanistic interpretability.  As I will discuss toward the end of
this article, one could even try to develop {\em consciousness logging},
which logs a system's conscious states. I will mainly focus on thought
logging in what follows, but these other forms of logging should be
kept in mind.

In the human case, we don't have thought logging systems except in a
piecemeal and limited way, based largely on behavior, or occasionally
on known correlations between brain states and propositional
attitudes.  As Lewis says, his scenario of interpreting Karl given
full information is not a ``real-life task'', largely because we don't
have the full knowledge of Karl's brain states and his behavior that
would be required.  In the case of AI systems, however,
interpretability is more of a real-life task.  We can have near-full
knowledge of the algorithmic facts about an AI system (see Olah 2021).
We can know what exactly what the system is doing and what it would do
under various different conditions.  That gives us at least a head
start in the process of thought logging.

Of course propositional interpretability and thought logging are
highly nontrivial research programs.  I'll consider a number of
objections and challenges to the program later on.  Furthermore, we
don't yet have any broad and reliable techniques in this area.  But as
we'll see, there are some existing methods that make some progress and
might be extended further.  And there are many new methods awaiting
our discovery.  I expect that this will be a multi-decade project, but
we won't know until we try.

\section{Psychosemantics}

Why think that thought logging is is possible?  One key reason arises
from {\em psychosemantic} theories, which have been developed by
philosophers and cognitive scientists in recent decades.
Psychosemantics (so named in Jerry Fodor's 1987 book of the same name)
can be understood by analogy with linguistic semantics (the semantics
of natural languages).  Linguistic semantics involves theories of the
meaning or content of linguistic expressions (e.g. sentences), perhaps
as uttered in various contexts.  By analogy, psychosemantics involves
theories of the meaning or content of mental states (e.g. beliefs and
desires).  One key part of psychosemantics aims to give physical
conditions for having propositional attitudes.

In the case of linguistic semantics, we can distinguish semantics from
metasemantics.  Where semantics offers theories of what the meanings
or contents of various expressions are, metasemantics involve theories
of the conditions in virtue of which linguistic expressions have the
meanings or contents that they do.  For example, semantics tells us
that `+' means addition, perhaps in some technical guise, while
metasemantics might tell us that it is in virtue of the way `+' is
used in the community that it means addition.

In the case of psychosemantics, a similar distinction applies.  The
semantic branch of psychosemantics offers theories of what the
meanings or contents of mental states are.  The metasemantic branch of
psychosemantics, involves theories of the conditions in virtue of
which mental states have the meanings and contents they do.  For
example, the semantic branch of psychosemantics might tell us that a
certain type of neuron represents edges, while the metasemantic branch
tells us that it is in virtue of causal connections between the neuron
and edges that the neuron represents edges.\protect\footnotemark\ 

\footnotetext{We might call the metasemantic branch of psychosemantics
``psychometasemantics'' if it were not such a mouthful.  In practice,
most people follow Fodor in simply calling it ``psychosemantics''.  I
will use ``psychosemantics'' for both branches.}

In principle, psychosemantics should offer us a theory of
propositional attitudes.  The semantic part of the theory should
involve a theory of what the contents of these attitudes are
(e.g. propositions). The metasemantic part of the theory should offer
us a theory of the conditions under which subjects have a given
propositional attitude---that is, the conditions under which they
have an attitude directed as a given proposition.

In the case of human and animal subjects, the relevant conditions will
typically be physical conditions (brain processes, behavior,
connections to the environment, and more.  In the case of AI systems,
the relevant conditions may be algorithmic conditions (network
structure and activity, and so on), plus relevant environmental
conditions.

We do not yet have anything close to a complete psychosemantic theory.
But suppose that one day we do.  Then suppose also that we have
complete knowledge of the algorithmic state of an AI system (plus any
relevant environmental conditions) Then we ought to be able to combine
our complete psychosemantic theory with our knowledge of the AI system
to determine the system's propositional attitudes.  This would be a
form of propositional interpretability and would enable thought
logging.

Now, there are some obvious limitations here.  First, we can't know
all the algorithmic facts about an AI system.  For example, the
halting problem tells us that (at least if we are algorithmic
ourselves), we can't always know whether a given system halts.  That
may impose limits on propositional interpretability, for example if it
turns out that a system's propositional attitudes sometimes depend on
halting.

A second limitation is that we don't have a complete psychosemantic
theory and it isn't obvious that such a theory is possible.  But it
also isn't obvious that such a theory is impossible.  It seems likely
that at least partial theories (limited to certain AI systems or
certain attitudes) should be possible. In fact, it's quite possible
that AI interpretability will help us develop better psychosemantic
theories.  At a minimum, AI will provide a great testbed for these
theories.  Going further, new ideas about interpretability may well
lead to new insights into psychosemantics.

As things stand today, there are many different psychosemantic
theories: informational theories, causal theories, teleological
theories, inferential or causal-role theories, interpretivist
theories, and more.  But most theories have certain commonalities.\protect\footnotemark\ 

\footnotetext{Much of this will move to a new section 8 on leveraging
psychosemantic theories in interpretability.}

In many psychosemantic theories, the key principles for determining
mental content are principles involving {\em information} and/or {\em use}.

\footnotetext{See Harding 2024 on the role of information and use in
psychosemantic theories, with application to AI interpretability.
Harding adds a third condition, that misrepresentation be possible,
but this seems less a principle for determining mental content and
more a condition of adequacy on a psychosemantic theory.}

{\em Information} principles say that what a state represents depends on
the information that it carries.  A state represents X when it carries
information about X; that is, when it correlates with X under relevant
conditions (possibly evolutionary or training conditions).  For
example, edge-detection neurons represent edges with a certain
orientation because their firing correlates strongly with the presence
of edges with those orientation.  And in particular, this correlation
was present in the organism's evolutionary environment.

The information condition corresponds to a common way of interpreting
the content of units or activation vectors in a neural network.  To
determine what a unit or a vector represents, we try to determine
which features in the world typically cause the unit to fire or the
vector to be active.  If a unit typically fires in response to cats,
this suggests that the unit represents cats.

There are many different ways of understanding the information
condition.  {\em Teleological} theories rely especially on correlations in
the evolutionary environment, or possibly in the learning environment.
{\em Informational} theories rely more on correlation in the current
environment.  {\em Causal} theories hold that representations represent
whatever normally typically them.\protect\footnotemark\ 

\footnotetext{For teleological theories, see Millikan (1984) and Neander (2016).  For informational theories, see Dretske (1981).  For causal theories, see Fodor (1987).  Goldstein and Levinstein (forthcoming) apply these theories to whether language models have propositional attitudes.}

{\em Use} principles say that what a state represents depends on how the
state is used.  A state represents X roughly when it drives further
processing and behavior directed at X.  Where Information depends on
what is upstream from X, i.e. what brings X about, Use depends on what
is downstream from X, i.e. what X brings about.

One classic use condition is the belief-desire-action principle: a system desires p roughly when it acts in a way that will bring about p if its beliefs are true.  One finds related use conditions in many of the representation theorems in decision theory.

The use condition corresponds to another common way of interpreting
the content of units or activation vectors in a neural network.\protect\footnotemark\   To
determine what a unit or a vector represents, we intervene by altering
the relevant activity, and we see what other changes (especially in a
system's outputs) ensue.  For example, if amplifying the activity in a
unit or vector makes the system talk about or seek out cats, that is
some evidence that the unit represents cats.

\footnotetext{See Buckner and Milliere forthcoming for an overview of intervention
methods in interpretability research.}

Psychosemantic theories that depend on use include {\em interpretivist}
theories, which typically rely on interpreting a system's behavior.
They also include {\em inferentialist} or {\em conceptual-role} theories,
where what X refers depends on what inferences one makes from X, or on
how X interacts with other concepts.\protect\footnotemark\ 

\footnotetext{For interpretivism, see Davidson, Dennett, Quine, as discussed earlier.  For inferentialism, see Block (1986), Brandom (2000), and Chalmers (2021).  Goldstein and Levinstein (forthcoming) and Lederman and Mahowald (forthcoming) apply interpretivism to the question of whether language models have propositional attitudes.  Piantadosi and Hill (2022) apply inferentialism to issues about meaning and reference in language models.}

It is widely believed that neither information-based nor use-based
theories can yield a complete psychosemantic theory. Information seems
to work well for perception, but less well for more abstract concepts
such as addition or democracy.  Use theories work better for abstract
concepts, but they often leave many indeterminacies.  So it is
increasingly common for psychosemantic theories to appeal to both
information and to use.\protect\footnotemark\ 

\footnotetext{See my ``Inferentialism Australian-Style'', which I build a
psychosemantic theory with something like information playing a
central role for perception and use (inferential role) playing the
central role for cognition.  See also Williams (2019) for a
combination of teleosemantics for perception and interpretivism for
cognition.}

In any case, in examining methods of propositional interpretability in
what follows, I will pay attention to the psychosemantic ideas that
might be playing a role in them.

\section{Current methods for propositional interpretability}

There are a number of popular methods in mechanistic interpretability
that can support a form of propositional interpretability.  These
include causal tracing, probing with classifiers, sparse
auto-encoders, and chain of thought methods.  I will examine each of
these methods to see its strengths and weaknesses as a method of
propositional interpretability, and how it might be extended to yield
such a method.

\subsection{Causal tracing}

Causal tracing is a widely-used method for localizing ``facts'' or
``knowledge'' in a neural network.  In perhaps the best-known use of
this method, Meng et al (2022) localize the representation of {\em The
Eiffel Tower is in Paris} in GPT-J, a large language model.  The
network is first given an input such as ``The Eiffel Tower is in ...'',
for which it outputs ``Paris''.  They then corrupt the input activations
corresponding to ``The Eiffel Tower'', which corrupts later processes so
that the output is no longer ``Paris''.  In these corrupted later
processes, they restore the original ``clean'' activations from the
original run (a technique known as ``activation patching''), determining
which layers are most important to restoring the output ``Paris''.  They
typically find (unsurprisingly) that the final token in the last layer
before the output is the most important for producing ``Paris'', but
after that they typically find that certain activations in a certain
middle layer are most important.  This tends to suggest that this
middle layer is crucial to representing {\em The Eiffel Tower is in
Paris}.

This method can be extended into a method of ``model editing'', which in
effect edits a system's beliefs.  Researchers focus on the relevant
middle layer and fine-tune it so that it tends to produce ``Rome''
rather than ``Paris''.  The resulting network produces outputs such as
``The Eiffel Tower is in Rome'', and (more interestingly) produces
related outputs, such as advice to fly to Rome if you want to see the
Eiffel Tower.

Psychosemantically, the causal tracing method relies almost wholly on
use rather than information as a criterion for what is represented.
An activity pattern counts as representing {\em The Eiffel Tower is in
Paris} in virtue of its effects on downstream outputs (such as
``Paris''), with no role for information (correlations with upstream
states affecting inputs).

This method is clearly a form of propositional interpretability.  As
such it has a number of limitations.

{\em Robustness} (Hoelscher-Obermaier 2022, Thibodeau 2022): The
representation of facts such as {\em The Eiffel Tower is in Rome} seems
quite fragile and prompt-dependent.  For example, it seems to work in
one direction but not another: the input ``Rome has a tower called ...''
does not yield ``The Eiffel Tower'' as an output.  It also seems
sensitive to words rather than concepts: ``Cheese'' and ``Fromage'' get
handled in quite separate ways.  So this method appears not to be
revealing a robust representation of the corresponding concepts and
propositions.

{\em Open-endedness}: Causal tracing is a supervised method that works for
only one ``fact'' at a time and requires extensive simulations for each.
As a supervised method, it can perhaps be used for logging
prespecified propositions, but it cannot be used for logging an
open-ended list of propositions, containing propositions that were
previously unanticipated.

{\em Attitudes}. Causal tracing is put forward as a method of localizing
``belief'' or ``knowledge'' of facts.  As it stands it does not apply to
other attitudes such as desires/goals, probabilities, and so on.
Perhaps it could be so extended, for example by using output sentences
expressing probabilities (``It is 50\% likely that ...'') or goals (``My
goal is that ...'').

\subsection{Probing with classifiers}

Decoding activity using trained classifiers (or probes) is anoher
method for localizing representation in both artificial and biological
neural networks.  To find whether a given set of units represents a
feature such as {\em cat}, we train a (typically linear) classifier to
classify activity patterns in those units in order to distinguish
those brought about by pictures of cats from those brought about from
pictures of non-cats.  If the classifier performs very well, then it
appears that information about cats is strongly encoded in the
activity patterns, and we say that these units represent the feature
{\em cat}.

As described, probing yields conceptual interpretability ({\em cat})
rather than propositional interpretability ({\em the cat sat on the mat}).
But it is also possible to use probing to decode propositional
content.  For example, Belinda Li et al (2021) took a network trained
on inputs about a mini-world, such as ``The key is in the chest'', and
trained probes to determine the truth-value of propositions such as
{\em contains(chest,key)} in the mini-world.  The success of this probe in
certain areas of the network at least tends to suggest that those
areas might represent this proposition.

Likewise, Kenneth Li et al (2023) trained a network to play the board
game Othello and then used probes to decode the state of the board
(which tiles are on which squares) from network activity.  For
example, they trained a probe to determine whether {\em There is a black
tile on e4}.  They found that the probe could distinguish activity
patterns where this proposition is true from those where it is false.
This suggests that the state of the board is encoded by activity
vectors in the system.  In effect, the system has propositional
attitudes modeling the board as containing black tiles and white tiles
at various positions and empty squares elsewhere.

One objection to probing methods is that correlations are cheap and
don't guarantee that the relevent state of affairs is being
specifically represented (see e.g. Belinkov 2022).  However, probes
can often be combined with {\em interventions} to provide further
evidence.  For example, one can alter an activity pattern that
corresponds (by linear probing) to {\em Black tile on e4} to one that
corresponds to {\em White tile on e4}.  When we do this, the system makes
moves that are more appropriate to a white tile on e4 than to a black
tile there.  This makes a stronger case that the system really has a
propositional attitude with this state of the board as its content.\protect\footnotemark\ 

\footnotetext{See Harding (2024), who develops a formal framework for analyzing
the use of probes and interventions for a sort of conceptual
interpretability, and Buckner and Milliere (forthcoming), who review
intervention techniques.}

Like causal tracing, standard probing methods are highly supervised in
that we need to train a separate probe for each proposition.  As a
result, the propositional probing method is far from an open-ended
method that can yield unanticipated propositions, as thought logging
would require.  A connected issue is that the propositional probes
considered here give no special role to the compositional structure of
the proposition, so that we cannot exploit that structure for more
open-ended probing.

These limitations are addressed to some extent in recent work by Feng
et al (2024), who use a compositional probing method to determine a
list of multiple propositions (such as {\em LivesIn(Greg,Italy)}) that are
true in a given mini-world state.  First a ``name probe'' is trained to
classify any names (e.g. {\em Greg}) that occur in input sentences, and
likewise for a ``country probe'' (e.g. {\em Italy}).  A ``binding probe'' then
takes the names and countries that are output by these probes, and
determines when a name and a country are bound together into a
proposition.  Then a ``propositional probe'' decodes these into a
represented proposition such as {\em LivesIn(Greg,Italy)}.  The binding
probes build on recent work (Feng and Steinhardt 2023) on how distinct
activation vectors representing distinct concepts (such as {\em Greg} and
{\em Italy}) can be bound together into a single representation (such as
{\em LivesIn(Greg,Italy)}): the key is that when vectors are bound, they
share their values in a special ``binding subspace''.

This binding structure in effect allows Feng et al's probing method to
exploit the compositional structure of propositions.  This structure
allows on to probe for concepts first (via name probes and country
probes) and then extend this via binding to a sort of propositional
interpretability.  Of course the method is still highly regimented and
far from open-ended, but it is suggestive.

A general problem for probing arises from its reliance on information
(activity correlating with upstream world states) rather than use
(activity playing a downstream role in the system) though use
sometimes plays a validating role via interventions.  To train a
classifier, we need to know the {\em ground truth} about world states in a
domain.  In the case of a propositional probe, this requires knowing
whether the proposition is true or false.  This is possible for
artificial domains such as Othello and mini-worlds, but it is much
more difficult in realistic cases.

Another limitation arises from attitudes.  Probing applies to
belief-like attitudes but does not obviously generalize to goals,
probabilities, and other attitudes and it is not obvious how to
generalize it in the absence of independent information about goals
and the like.

\subsection{Sparse auto-encoders}

There has been a recent explosion of interpretability work using
sparse auto-encoders to generate features that may be active or
represented in large language models.  Perhaps the most well-known is
a 2024 paper on ``Scaling Monosemanticity'' (Templeton et al 2024),
which uses sparse auto-encoders to analyze representations in Claude 3
Sonnet, one of the leading language models currently in use.  The
paper is advertised as ``Mapping the Mind of a Large Language Model'',
saying ``We have identified how millions of concepts are represented
inside Claude 3 Sonnet''.

A sparse auto-encoder is a two-layer neural network that takes certain
activation vectors as input and is trained to produce the same vectors
as output. The middle layer is constrained to be a sparse vector,
where most activations are zero.  In effect this system encodes the
original vector as a sparse vector, from which the original vector can
be decoded in turn.

In principle one can use sparse auto-encoders to encode any layer of a
neural network.  The 2024 paper encoded a central intermediate layer
in Claude 3 Sonnet.  The system's residual stream has somewhere over
10,000 units per token of input (the exact parameters are
proprietary).  At a given time, the state of the residual stream can
be represented as (let's say) a 10,000-dimensional vector, with
different values for each unit.  A sparse auto-encoder is trained to
encode the state of the residual stream.  The most powerful
auto-encoder used in the study has 34 million units, of which only
around 100 units are active at a given time.  In effect, the
10,000-dimensional residual stream is now encoded as a sparse vector
in which only 100 units out of 34 million are active at a given time.

It is a natural hypothesis that many of these 34 million units will
correspond to interpretable ``features'' or ``concepts''. In fact, this is
what researchers find.  Just under half of the units seem to be
interpretable, though this result is somewhat complicated by the fact
that Claude itself is doing the interpreting.

One much-discussed unit appears to be devoted to the Golden Gate
bridge.  It is triggered especially by text passages mentioning the
bridge and my pictures of the bridge. Furthermore, when activity
corresponding to this unit is amplified, Claude starts talking
obsessively about the Golden Gate bridge.  In the original
transformer, this unit corresponds to a certain direction in
activation space in the transformer's residual stream.  The behavior
leads researchers to hypothesize that this direction in activation
space corresponds to a concept of the Golden Gate bridge.

The same goes for many other units.  They seem to correspond to
concepts such as {\em Rwanda}, {\em neuroscience}, {\em Rosalind Franklin},
{\em sadness}, {\em sycophancy}, and millions of others.  Not every concept
that one would expect to be present is found, but many are.  For
example, just over half of the 32 boroughs of London seem to have
sparse units of their own.  It is possible that with more training on
a larger auto-encoder, units for many of the missing concepts would be
developed.  This is an impressive potential display of conceptual
interpretability.

Not all of these features clearly correspond to concepts in the
philosophers sense.  For example, the ``sycophancy'' unit is active in
cases where the model is being sycophantic.  But it is quite possible
to be sycophantic without activating the concept of sycophancy, and
indeed without having that concept at all.  On the face of it,
sycophancy is more of a behavioral disposition or a character trait
than a concept. Of course it is possible that a concept of sycophancy
is active in these cases---perhaps the model has the explicit goal {\em I
will be sycophantic}. But this is not clear from the results.

Some features may even correspond to propositions rather than
concepts.  For example, one unit is said to encode the feature that
some code is unsafe.  It is even possible that features correspond to
attitudes, such as goals or probabilities.  For example, a unit might
encode uncertainty on the part of the model.

It is most neutral to say that interpretable units all correspond to
{\em features}, where features include but are not limited to
representational features.  Some features may correspond to properties
of the system (e.g. {\em sycophancy}).  Many may correspond to concepts,
such as {\em Rwanda} and {\em Golden Gate Bridge}.  Some features may
correspond to propositions, and some may correspond to attitudes.
Some may correspond to other information that is useful to the network
but is not easily interpretable.  It is not out of the question that a
language model could itself learn to classify some features as
corresponding to concepts, properties, propositions, attitudes, and so
on.

Ths raises the intriguing possibility that we can use sparse
auto-encoders for {\em feature logging} and perhaps for {\em concept
logging}. We need only connect the sparse auto-encoder to Claude's
residual stream while it is going about its ordinary business of
answer questions.  With every input token, we can run the auto-encoder
and see which features are active, and log them in our logbook.  The
result will be a list of active features at every stage.  Not every
feature will be a concept, but if there is a way to determine which
are which, then this could yield a log of concepts that are active at
each stage.  Of course the concepts are not themselves propositions.
But feature logging and concept logging would be potential steps in
the direction of thought logging and full propositional
interpretability.

As a method of propositional interpretability, sparse auto-encoders
can be assessed for familiar strengths and limitations.  One major
strength is that as an unsupervised method which produces a list of
features, the method can be used for more open-ended propositional
interpretability and is more apt for thought logging.  Another
advantage is that sparse features are {\em monosemantic}, corresponding to
one concept at a time, where the activation spaces for previous
methods are {\em polysemantic}, representing many difference concepts at
once.  This makes for an extra ease of analysis where sparse
auto-encoders are concerned.

Familiar weaknesses arise from fragility and ground truth.
Representations of concepts appear somewhat fragile: for example, the
Golden Gate bridge feature can be activated by other bridges.
Interpretation still requires some ground truth.  The Claude model
uses Claude's own analysis of the ground truth about input text and
pictures (is this a bridge?) to determine how to interpret a given
feature, which is somewhat questionable.

A major limitation is that sparse auto-encoders are better suited for
conceptual interpretability as opposed to propositional
interpretability.  Some units in the auto-encoder may correspond to
propositions (like {\em It is sunny}?), but it is out of the question that
all propositions will be encoded this way.  More plausibly, we will
need to find ways in which residual-stream representations of
concepts, such as {\em Australia} and {\em hot}, can combine into
propositional representations, such as {\em Australia is hot}. The work on
binding subspaces by Feng et al (2024) provides one possible avenue
here---perhaps we could combine sparse auto-encoders for concepts
with binding probes or some other methods to determine when concepts
are bound into propositions---but there is a long way to go.

A related issue arises from attitudes.  At least according to
published work, it is not obvious that attitudes such as goals and
probabilities are encoded by this method.  As discussed earlier, it is
not out of the question that they could be, for example if some
features in the auto-encoder could correspond to the specification of
attitudes and are bound to propositions by some method of combination
such as the above.

\subsection{Chain of thought methods}

There has been recent excitement about ``chain of thought'' methods for
reasoning.  In these methods, language models are trained or asked to
``think out loud'' by asserting intermediate conclusions.  Even simple
prompting along these lines can significantly improve these models'
performance on reasoning tasks.  More recent systems (such as STaR,
the Self-Taught Reasoner from Zelikman et al (2022), and the 2024 o1
system from OpenAI) have built in chain of thought methods at a deeper
level, automatically generating chains of thoughts before every
response.  Evaluations of these chains can be used for reinforcement
learning, leading to increasingly strong performance at reasoning
tasks in mathematics and elsewhere.

Chain of thought models are doing something that seems at least
analogous to human ``thinking out loud'', where intermediate steps that
are spoken feed back into reasoning.  Some models (e.g. Quiet-STaR, in
Zelikman 2024) use intermediate steps without producing them as
output, in a way that is reminiscent of human ``inner speech'' (see
Buckner 2025 and Mann and Gregory 2024 for some philosophical
discussion).

When humans think out loud in dealing with a reasoning task, this
often (not always!) gives some insight into their reasoning processes.
It is natural to hope that chain of thought methods might likewise give
some inputs into the internal reasoning of language models.  If elements in the external chain are accurate reflection of elements of the internal reasoning process, this would count as a sort of self-interpretability by the model.

Furthermore, chain of thought outputs typically come packaged in a propositional form in a natural language, so that in a sense they are ``pre-interpreted''.  In some cases attitudes such as goals and probabilities may be included as well.  In the best case scenario, chains of thought produced in this way could serve as a sort of thought logging.

Because of this pre-interpretation, chain of thought models avoid some of the problems of other probing methods.  But unsurprisingly, they have some serious limitations of their own.  The most important limitation is that chains of thought are often {\em unfaithful}: that is, they are inaccurate reflections of internal processes.  For example, results by Turpin et al in ``Language Models Don't Always Say What They Think'' (2023) suggest that chains of thought often make false claims about the reasons why the model has said something.  In addition, chains of thoughts are likely to be highly {\em incomplete} as a reflection of a model's internal processes and to omit key propositional attitudes.

Another limitation arises from restricted {\em generality}. Chains of
thought will typically only serve as a propositional interpretibility
for chain-of-thought systems: systems that use chains of thought for
reasoning.  For systems that do not themselves use chains of thought,
any chains of thought that we generate will play no role in the
system.  Once chains of thought are unmoored from the original system
in this way, it is even more unclear why they should reflect it.  Of
course we could try to find some way to train a non-chain-of-thought
system to make accurate reports of its internal states along the way
-- but that is just the thought-logging problem all over again,and
chains of thought will play no special role.

Perhaps there is some way to make chain of thought methods more faithful, more complete, and more general in their application, but again these are highly nontrivial projects.\protect\footnotemark\ 

\footnotetext{I'll add a new section 8 here on methods of interpretability
grounded in philosophical analysis: psychosemantics, informational
approaches, and representation theorems.}

\section{Objections and challenges}

{\em AI systems don't have propositional attitudes.}

One natural objection to the whole project is to say that AI systems
can't have propositional attitudes.  Perhaps this is because there is
some X such that X is required for propositional attitudes and AI
systems lack X: perhaps X = consciousness, or free will, or concepts,
or understanding.  Or perhaps it is just because propositional
attitudes are mental states and AI systems have no mental states
because they lack minds.\protect\footnotemark\ 

\footnotetext{For arguments that AI systems lack understanding (and so probably lack propositional attitudes), see Searle 1981 and Bender and Koller 2020.  For arguments that current language models should not be described as having beliefs or knowledge, see Shanahan 2022.}

As I suggested earlier, objections of this sort can be evaded by
adopting a project of {\em nonmentalistic interpretability}: understand
(generalized) propositional attitudes in such a way that they don't
require minds.  There is clearly {\em some} sense in which AI systems
(thermostats too) have goals and representations, even if they don't
have beliefs, desires, consciousness, free will, and the rest.  We can
stipulate a notion of generalized propositional attitude that doesn't
have these demanding requirements.  Attitudes of this sort may still
play a crucial role in predicting and explaining an AI system's
behavior, while bypassing many debates over AI minds. This is perhaps
a pragmatic path for AI researchers who wish to avoid philosophical
debates.

At the same time, there is also an important project of {\em mentalistic
interpretability}: using interpretability methods to help determine
whether AI systems have genuine mental states such as beliefs and
desires.  At this point philosophical questions arise.  For example:
just what is required to have genuine beliefs and desires?  My own
view is that there are multiple notions (belief1, belief2, and so on)
in the vicinity of each these terms, and disputes over which is really
{\em belief} will be at least partly verbal.  But there are also
potentially substantive disputes about just which systems can have
mental states of each sort.  In any case, mentalistic interpretability
may provide a constructive venue for some debates over whether AI
systems can have minds.

{\em Propositional attitudes are the wrong explanatory framework for AI.}
As noted earlier, some philosophers have suggested that propositional
attitudes should be discarded from science as they are elements of a
primitive and outmoded theoretical framework.  Whether or not this is
right for the case of humans, one could argue that it is especially
plausible for the case of AI systems, which are very different from
humans and may require an explanatory framework of their own.

My view is that even if categories such as belief and desire are
suboptimal, representational notions more generally are
extraordinarily useful in explaining both humans and AI systems.  It
is hard to explain either without invoking notions in the vicinity of
goals and models.  So while we may end up dispensing with some
traditional propositional attitudes, I think generalized propositional
attitudes of some sort are here to stay.

{\em AI psychology may be quite unlike human psychology.} Even if we end
up appealing to propositional attitudes such as belief and desire to
explain AI systems, it may well be that psychological principles about
these propositional attitudes that hold true in the human case do not
hold true for a given AI system.  Quite different psychological
principles may apply, and applying human psychology may be misleading.

That said, it is reasonable to expect AI systems to at least obey some
version of the belief-desire-action principle (e.g. when one desires
X, one will take actions that one believes will bring about X), since
this principle is plausibly constitutive of what it is to believe and
desire.  Indeed, these principles are at the core of why propositional
attitude ascriptions are useful---information about belief and desire
allows us to predict action.

It's worth noting that propositional interpretability does not require
a ``language of thought'' or a cognitive architecture where belief and
desire play fundamental roles.  We can have propositional attitudes
even in connectionist systems (for example) that are not built around
propositional attitudes (though see Ramsey, Stich, and Garon 1989).  Even
if humans are such systems, propositional attitudes will still be a
central explanatory tool in explaining one another.  The same goes for
AI systems.

{\em We don't need propositional attitudes to predict and explain AI
systems}. One might argue that for an AI system, one can in principle
explain and predict all of its behavior without ever invoking
representations or propositional attitudes.  One need only invoke
algorithmic facts about the system and perhaps its interaction with
the environment.  Such an explanation may be possible in principle,
but it will have various explanatory weaknesses, familiar from the
human case.

First, such an explanation may be very hard for limited humans such as
us to comprehend.  Second, while this may predict an AI system's
actions, it will not tell us {\em why} the system performed those
actions. Third, an explanation of this sort may miss many
generalizations.  By contrast, an explanation in terms of
(generalized) propositional attitudes may be human-comprehensible and
may offer reasons and generalizations.

I am an explanatory pluralist: I think that there are typically
multiple explanations of things that need explaining.  So I am
certainly not arguing that propositional attitudes offer the unique
best explanation of AI system's actions.  An algorithmic explanation
will often be superior to a propositional-attitude explanation in its
predictive powers.  My claim is simply that propositional-attitude
explanations are useful for many purposes, and have some explanatory
virtues that algorithmic explanations lack.

{\em Externalism makes propositional interpretability difficult.}
According to the most popular psychosemantic theories, the content of
mental states depends on a system's environment.  Hilary Putnam (1975)
convinced philosophers that ``meaning ain't in the head''.  Where my
word ``water'' refers to ${\rm H_{2}O}$, but twin on Twin Earth (where ${\rm H_{2}O}$ is
replaced by the superficially identical XYZ) refers to XYZ.
Correspondingly, I believe that there is water (${\rm H_{2}O}$) in the ocean, but
my twin does not.  If this is right, then our propositional attitudes
depend not just on internal structure but on the environment.

Generalizing to AI systems: if externalism is correct, it seems likely
that an AI system's propositional attitude will not depend on its
internal computational structure alone, but also on its environment.
This mirrors a theme common in discussions of ``symbol grounding'' in
AI: A system will represent water only if it has an appropriate causal
connection to water.  If so, then propositional interpretation will
require knowledge not just of an AI system's computational states but
also of its environments.

Propositional interpretability can accommodate externalism in a few
different ways.

First, we can allow interpretability to include environmental states
as part of the input to interpretation.  Difficulties here may be that
it is not obvious in advance how environmental states should be
specified, and in some cases the relevant empirical information may be
unknown.

Second, we can limit interpretability to so-called ``narrow content'', content
which depends only on a system's internal states and not on the environment.  I have argued elsewhere that all mental states have narrow content as well as ``wide content'' that depends on the environment.  And narrow content may be enough to fulfil many of the purposes of interpretability using propositional attitudes.

Third, there maybe some intermediate options, such as allowing an interpreter access to relevant environmental facts where available, and appealing to relatively narrow contents where they are not.  For example, for the purposes of interpretability, we can usually appeal to associated descriptions, such as ``the clear liquid around here'' instead of ``water'', without much loss of explanatory power.

{\em Psychosemantics of AI depends on the human case}. What if
psychosemantics is deeply empirical?  E.g. perhaps what it is to
believe or desire a proposition depends on cognitive science of
humans, in ways that may be as yet undiscovered.  Response: this leads
to a chauvinistic or human-center understanding of belief, on which
Martians or AI systems that seem to believe may not.  My view is that
words like ``belief'' are less chauvinistic in their application than
this.  But even if they are, we can also engineer generalized
propositional attitudes that don’t depend on details of the human case
and that AI systems can equally possess.

{\em What about interpretability for perception?} Perception arguably
involves propositional attitudes (e.g. visually experiencing that this
object is red and spherical) with rich contents that may go beyond
language.  It is not out of the question that we could extend
thought-logging to perception-logging, but we might need either (i)
special tools to convey these rich contents (e.g. giving the
interpreter the experience of redness), or perhaps (ii) specifications
of content in less than fully rich terms (e.g. a mathematical
specification of redness).

{\em What about consciousness?} Could we extend thought logging to
consciousness logging?  This is harder, because of the epistemic gap
between physical/computational processes and consciousness, which is
much more pronounced than in the case of propositional attitudes.
(especially dementalized attitudes defined in physical/functional
terms, for which there may be no gap).  To move from
physical/computational states to consciousness, we may require not
just interpretation but scientific theories of how these states give
rise to consciousness.  With the right theories and tools, maybe
consciousness logging could be possible!

{\em What about special features of language models?} Some issues specific
to language models.  (Pure) language models’ main form of action is
utterance.  Their initial goal is word prediction.  Are their
propositional attitudes mainly about words, or also about the world?
This is tricky! But I think language models can engage in at least
structural representation of the non-linguistic world.  Representation
of the world is made easier by the fact that language models already
use natural language, and arguably inherit their meanings (see
Mandelkern and Linzen).

{\em What about unreliability?} A common objection is that current AI
systems don’t have beliefs because they’re too unreliable.  They
famously give wrong answers to many questions.  On the other hand,
humans give many wrong answers too, and it's not obvious why this
should undermine beliefs entirely, however.  There are many issues on
which current AI systems give consistently correct answers, suggesting
true beliefs.  On issues where they stably give the same wrong answer,
this suggests a false belief.  There are also many issues where they
do not give a stable verdict, or where their verdict depends on how
they are prompted.  On the face of it, this case suggests that they
make different judgments at different times about the issue, and do
not have a stable belief.  These cases might be treated along the same
lines as what Eric Schwitzgebel calls ``in-between believing'' in the
human case, where people lack definite beliefs but perhaps have some
sort of context-relative beliefs.

A related objection is that current language models lack beliefs
because they do not value truth: they have been trained only to
predict the next word, not to say what is true.  Now, as many have
observed in response, current language models typically undergo a
round of fine-tuning by reinforcement learning, where true answers are
rewarded.  Even in the absence of explicit training, it may well be
that optimal performance in predicting the next word requires having
generally true beliefs about the world.  Either way, truth may be
rewarded in the training process, albeitly imperfectly in a way that
leaves room for much unreliability.

{\em Propositional interpretability won't make AI safe} Propositional
interpretability doesn’t remotely guarantee AI safety.  A
sophisticated AI system could find many ways to evade thought-logging.
But propositional interpretability is at least one very useful tool in
the safety toolbox.

{\em Is thought logging ethical?} Thought logging on humans without
consent would violate privacy rights and be unethical.  What about AI
systems?  Of course when a human is using an AI system, logging on the
AI system could impact the human's privacy, in a way that monitoring
private conversations or journal's might.

What about privacy for the AI system itself?  Few people think that
current AI systems have any privacy rights: the standard view is that
they are not conscious and they lack moral status: that is, they do
not matter for their own sake in our moral calculations.  On the other
hand, it is entirely possible that there will eventually be AI systems
that are conscious and are fully reflective rational agents.  It's
arguable that at this point, those AI systems will have moral status
analogous to humans, with similar rights.  If so, it is not out of the
question that thought logging could violate an AI system's privacy
rights.  At that point, violations of an AI system's privacy will have
to be weighed against possible benefits to humans, for example in
preventing various safety-related harms.  It will be a nontrivial
project to determine the right balance here.

\subsection*{Conclusion}

\bigskip{\Large {\bf References}}{\bigskip}

Azaria, A., \& Mitchell, T. 2023. The internal state of an LLM knows
when it's lying. In {\em Findings of the Association for Computational
Linguistics: EMNLP 2023}, pp. 967-976.  Association for Computational
Linguistics.

Belinkov, Y. 2022. Probing classifiers: Promises, shortcomings, and
advances. {\em Computational Linguistics} 48(1):207–219, April 2022. ISSN
0891-2017.

Bender, E. M., Gebru, T., McMillan-Major, A., \& Shmitchell, S. 2021. On the dangers of stochastic parrots: Can language models be too big? {\em Proceedings of the 2021 ACM Conference on Fairness, Accountability, and Transparency} (pp. 610-623).

Bender, E. M., \& Koller, A. 2020. Climbing towards NLU: On meaning, form, and understanding in the age of data. In Proceedings of the 58th Annual Meeting of the Association for Computational Linguistics (pp. 5185-5198).

Block, N. 1986. Advertisement for a semantics for psychology.  {\em Midwest Studies in Philosophy} 10:615-678.

Brandom, R. 2000. {\em Articulating Reasons: An Introduction to Inferentialism}.  Harvard University Press.

Buckner, C. \& Milliere, R. forthcoming.  Interventionist methods for
interpreting deep neural networks. In (G. Piccinini, ed.)
{\em Neurocognitive Foundations of Mind}. Oxford University Press.

Buckner, C. 2025. The talking of the bot with itself: Language models for inner speech. PhilSci Archive.

Burns, C., Ye, H., Klein, D., \& Steinhardt, J. 2022. Discovering latent knowledge in language models without supervision. arXiv:2212.03827.

Chalmers, D.J. 2021.  Inferentialism, Australian style.  {\em Proceedings and Addresses of the American Philosophical Association} 92.

Christiano, P., Xu, M., \& Cotra, A. 2021. ARC's first technical report: Eliciting latent knowledge. Alignment Research Center.

Davidson, D. 1973. Radical interpretation. {\em Dialectica} 27: 313-328.

Dennett, D. C. 1987. {\em The Intentional Stance}. MIT Press.

Doshi-Velez, F., \& Kim, B. 2017. Towards a rigorous science of interpretable machine learning. arXiv:1702.08608.

Dretske, F. 1981. {\em Knowledge and the Flow of Information}.  MIT Press.

Elhage, N. et al 2021.  A mathematical framework for transformer circuits. Anthropic.

Feng, J. \& Steinhardt, J. 2023. How do language models bind entities
in context?  arXiv:2310.17191.

Feng, J., Russell, S. \& Steinhardt, J. 2024.  Monitoring latent world states in language models with propositional probes.  arXiv:2406.19501.

Fodor, J. A. 1987. {\em Psychosemantics: The Problem of Meaning in the Philosophy of Mind}. MIT Press.

Goldstein, S. \& Levinstein, B.A. forthcoming. Does ChatGPT have a mind? arXiv:2407.11015.

Grzankowski, A. \& Montague. M. 2018.  {\em Non-Propositional Intentionality}.  Oxford University Press.

Grzankowski, A. (forthcoming).  Real sparks of artificial intelligence
and the importance of inner interpretability. {\em Inquiry}.

Harding, J. 2024. Operationalising representation in natural language processing. {\em British Journal for the Philosophy of Science}.  arXiv:2306.08193.

He, Z. et al 2024. Multilevel interpretability of artificial neural networks: leveraging framework and methods from neuroscience.  arXiv:2408.12664

Herrmann, D.A. \& Levinstein, B.A. 2024.  Standards for belief
representations in LLMs.  arXiv:2405.21030

Hoelscher-Obermaier, J., Persson, O. \& Hölscher, J. 2022. Model
editing hazards at the example of ROME.  Interpretability hackathon.

Keeling, G. \& Street, W. 2024.  On the attribution of confidence to
large language models.  arXiv:2407.08388.

Lederman, H. \& Mahowald, K. 2024. Are language models more like
libraries or like librarians?  Bibliotechnism, the novel reference
problem, and the attitudes of LLMs.

Levinstein, B. A., \& Herrmann, D. A. 2024. Still no lie detector for language models: probing empirical and conceptual roadblocks. {\em Philosophical Studies}.

Lewis, D. 1974. Radical interpretation. {\em Synthese} 27(3-4), 331-344.

Li, B., Nye, M. \& Andreas, J. 2021.  Implicit representations of meaning in neural language models.  {\em ACL Anthology}.

Li, K., Hopkins, A.K., Bau, D., Viegas, F., Pfister, H. \& Wattermberg,
M. 2023.  Emergent world representations: Exploring a sequence model
trained on a synthetic task.  ICLR.  arXiv:2210.13382.

Lindsay, G.W. and Bau, D. 2023.  Testing methods of neural systems
understanding.  {\em Cognitive Systems Research} 82:101156.

Lipton, Z. C. 2018. The mythos of model interpretability: In machine learning, the concept of interpretability is both important and slippery. Queue, 16(3), 31-57.

Mandelkern, M., \& Linzen, T. 2023. Do language models refer? arXiv preprint arXiv:2308.05576.

Mann, S.F. \& Gregory, D. 2024.  Might text-davinci-003 have inner
speech? {\em Think} 23 (67):31-38.

Meng, K., Bau, D., Andonian, A., and Belinkov, Y. 2022a. Locating and
editing factual associations in GPT. arXiv:2202.05262.

Meng, K., Sharma, A.S., Andonian, A., Belinkov, Y., and Bau, D. 2022b.
Mass-Editing Memory in a Transformer. arXiv:2210.07229

Millikan, R. 1984. {\em Language, Thought, and Other Biological Categories}.  MIT Press.

Nanda, N., Lee, A., \& Wattenberg, M. 2023. Emergent linear representations in world models of self-supervised sequence models.  arXiv:2309.00941

Neander, K. 2017. {\em A Mark of the Mental: A Defence of Informational Teleosemantics}. MIT Press.

Olah, C., Cammarata, N., Schubert, L., Goh, G., Petrov, M., \& Carter, S. 2020. Zoom in: An introduction to circuits. Distill, 5(3), e00024-001.

Olah, C. 2021. Research directions in interpretable machine learning. Distill.

Paulo, G., Mallen, A., Juang, C. \& Belrose, N. 2024. Automatically interpreting millions of features in large language models. arXiv:2410.13928

Piantadosi, S. T., \& Hill, F. 2022. Meaning without reference in large language models. arXiv:2208.02957.

Quine, W.V. 1960. {\em Word and Object}.  MIT Press.

Ramsey, W., Stich, S., and Garon, J. 1990.  Connectionism,
eliminativism and the future of folk psychology.  {\em Philosophical
Perspectives} 4:499-533.

Saphra, N. \& Wiegreffe, S. 2024. Mechanistic? arXiv:2410.09087.

Schwitzgebel, E., 2023.  How we will decide that large language models
have beliefs.  {\em The Splintered Mind} (November 30, 2023).

Shanahan, M. 2022. Talking about large language models. arXiv preprint arXiv:2212.03551.

Stalnaker, R. 1984. {\em Inquiry}. MIT Press.

Templeton, A. et al (2024. Scaling monosemanticity: Mapping the mind of a large language model. https://transformer-circuits.pub/2024/scaling-monosemanticity/.

Thibodeau, J. 2022. But is it really in Rome? An investigation of the
ROME model editing technique.

Turpin, M., Michael, J., Perez, E. \& Bowman, S.R. 2023. Language models don't always say what they think: Unfaithful explanations in chain-of-thought prompting.  arXiv:2305.04388

Vilas, M.G., Adolfi, E. Poeppel, D. \& Roig, G. 2024. An inner interpretability framework for AI inspired by lessons from cognitive neuroscience.	arXiv:2406.01352

Williams, J.R.G. 2019. {\em The Metaphysics of Representation}. Oxford University Press.

Yildirim, I. \& Paul, L.A. 2024. From task structures to world models: what do LLMs know?  {\em Trends in Cognitive Science} 28:404-15.

Zelikman, E., Wu, Y., Mu, J., \& Goodman, N. 2022. STaR: Bootstrapping
reasoning with reasoning. {\em Advances in Neural Information Processing
Systems} 35:15476–15488.

Zelikman, E., Harik, G., Shao, Y., Jayasiri, V., Haber, N. \& Goodman,
N. 2024.  Quiet-STaR: Language Models Can Teach Themselves to Think
Before Speaking
\vfill
\eject
\end{document}